\documentclass[11pt]{article}
\usepackage[preprint]{acl}
\usepackage{times}
\usepackage{latexsym}
\usepackage[T1]{fontenc}
\usepackage[utf8]{inputenc}
\usepackage{microtype}
\usepackage{inconsolata}
\usepackage{graphicx}
\usepackage{subcaption}
\usepackage{booktabs}
\usepackage{flushend}
\usepackage{float}
\usepackage{algorithm}
\usepackage{longtable}
\usepackage{placeins}
\usepackage{afterpage}
\usepackage{amsmath,amssymb}
\usepackage{array}
\usepackage{tikz}
\usetikzlibrary{arrows.meta,positioning,calc}
\hypersetup{pdftitle={FastHarness: Experience-Driven Review Scheduling for Vision-Language-Action Policies},pdfauthor={Tao Lin, Yuxin Du, Haoxiao Wang, Zile Zhong, Bo Zhao},pdfsubject={Technical report}}
\definecolor{BlogTeal}{HTML}{176F63}
\hypersetup{colorlinks=true,citecolor=BlogTeal,linkcolor=BlogTeal,urlcolor=BlogTeal}
\newcommand{\fh}{FastHarness}
\newcommand{\pizero}{\ensuremath{\pi_{0.5}}}
\newcommand{\MeanTokenReduction}{57.2\%}
\newcommand{\HybridMeanSuccessRate}{80\%}
\newcommand{\SharedTaskMeanSuccessRate}{97.19\%}
\title{FastHarness: Experience-Driven Review Scheduling\\for Vision-Language-Action Policies}
\author{
  Tao Lin$^{1,*}$ \quad
  Yuxin Du$^{1,*}$ \quad
  Haoxiao Wang$^{2,*}$ \quad
  Zile Zhong$^{1,*}$ \quad
  Bo Zhao$^{1,\dagger}$ \\[0.5em]
  $^{1}$School of AI, Shanghai Jiao Tong University \quad
  $^{2}$Zhejiang University \\[0.3em]
   taolin200108@gmail.com, bo.zhao@sjtu.edu.cn \\
  {\small $^{*}$Equal contribution. \quad
  $^{\dagger}$Corresponding author.} \\[0.5em]
  {\small
  \href{https://github.com/FastHarness/fast-harness}
       {https://github.com/FastHarness/fast-harness}}
}
\begin{document}
\raggedend
\raggedbottom
\maketitle
\begin{abstract}
Agentic robot systems combine language-model reasoning with robot policies to monitor execution and recover from failures. However, reviewing every action chunk incurs substantial overhead, even in familiar situations. We introduce \fh{}, an experience-driven execution harness that adaptively schedules GPT-6 Astra supervision of a \pizero{} policy. A fast path uses stage memory and reviewed outcomes to reduce review frequency as experience accumulates, while a complementary recovery path focuses supervision on correcting errors. Once recovery is verified, routine execution resumes, with failure experience retained to guide future interventions. The policy remains frozen and generates fresh actions from current observations. Across four simulated manipulation tasks, reviewer token use decreases by an average of \MeanTokenReduction{} per task from the first evaluated episode to the mean of the final five episodes. Mean episode time decreases from 85.5 to 52.0 minutes over the same interval. On the same tasks, mean success increases from \HybridMeanSuccessRate{} for GPT-as-Policy (\pizero{} + GPT-6 Astra) to \SharedTaskMeanSuccessRate{} with \fh{}. These results highlight adaptive supervision as a practical route to more efficient agentic robotics without additional model training.
\end{abstract}

\begin{figure*}[!t]
\centering
\includegraphics[width=\textwidth]{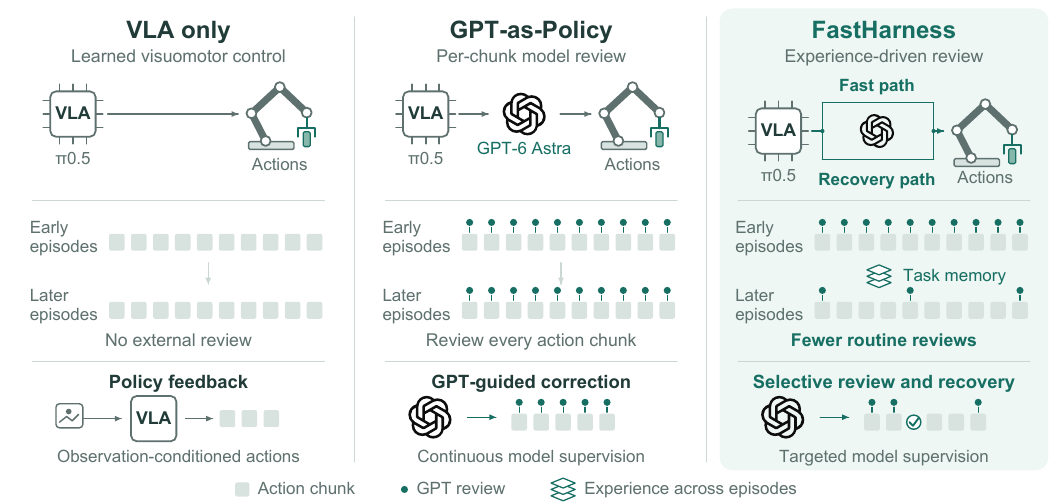}
\caption{From VLA execution to experience-driven supervision. VLA-only execution generates actions from current observations. GPT-as-Policy adds GPT-6 Astra review at every action chunk. Using the same policy and reviewer, \fh{} skips redundant reviews as task experience accumulates and concentrates supervision on error recovery. Dots denote GPT reviews; the check mark indicates verified recovery, after which adaptive review resumes.}
\label{fig:teaser}
\end{figure*}

\section{Introduction}

Agentic robot systems \citep{zhang2026harness,fu2026cap,lu2026aspire,shi2025hi} combine learned policies with language-model reasoning to monitor execution and recover from errors. A vision-language-action (VLA) policy \citep{kim2024openvla,black2024pi_0,lin2025evo,lin2026evo,lin2026evodepth,brohan2023rt,lin2026la4vla} maps observations and instructions to actions, while a language model uses execution feedback to assess progress and guide corrections \citep{huang2022inner}. GPT-as-Policy follows this approach by pairing \pizero{} \citep{intelligence2025pi_} with GPT-6 Astra \citep{su2026astra}. An execution harness coordinates the two models, determining when to request a review and how the resulting feedback affects subsequent actions.

Frequent model calls, however, can make execution slow and costly. Reviewing every action chunk repeats this overhead even in task stages that the policy has already completed reliably. Errors require a different response: a missed grasp or dropped object may need several rounds of correction before execution can continue. Repeated task execution provides evidence for distinguishing these situations. The challenge is to use that evidence to reduce redundant review while retaining the supervision needed for recovery.

We introduce \fh{}, an experience-driven harness that uses reviewed outcomes to decide when model supervision is needed (Figure~\ref{fig:teaser}). Repeatedly successful stages can proceed with less frequent review, while remembered failures prompt closer inspection when similar situations recur. The harness updates its memory and review schedule as experience accumulates. The \pizero{} policy remains fixed and generates each action chunk from the current observation.

\begin{figure*}[!t]
\centering
\includegraphics[width=.95\textwidth]{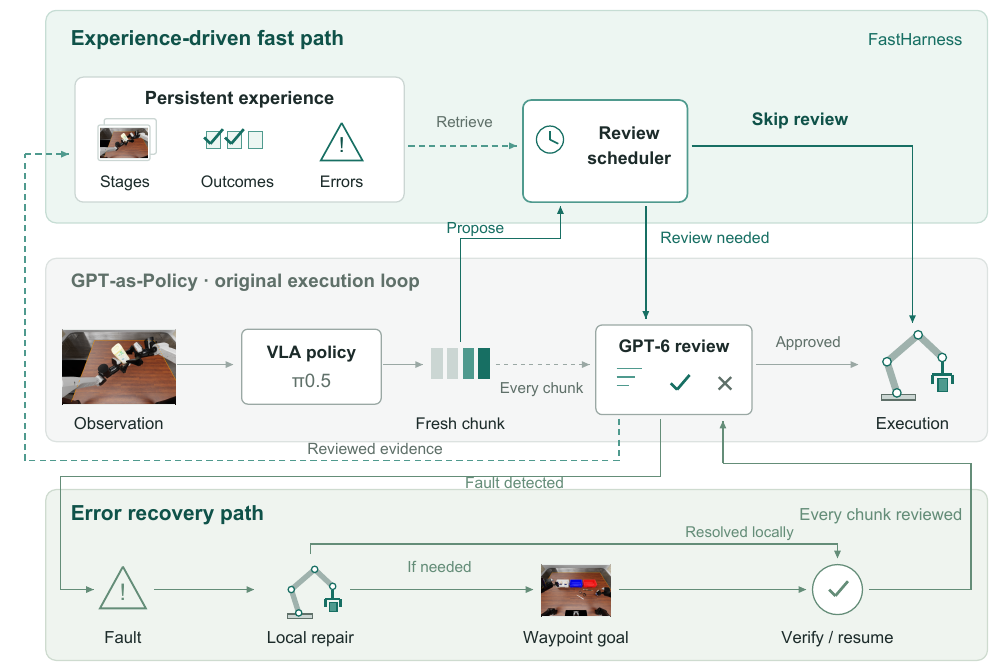}
\caption{\fh{} adds two paths around the original GPT-as-Policy execution backbone (gray, middle). The experience-driven fast path (top) routes fresh VLA proposals through a memory-guided scheduler, bypassing GPT-6 review when eligible; the gray dashed link denotes the original per-chunk review route. The recovery path (bottom) uses reviewed local repair and, when needed, verified waypoint goals. Ordinary scheduling resumes after verified resolution. Dashed teal feedback retains reviewed evidence across episodes. Image inserts illustrate recorded task scenes.}
\label{fig:system}
\end{figure*}

\fh{} implements this adaptation through two complementary paths (Figure~\ref{fig:system}). During routine execution, the \emph{fast path} uses the current task stage and its reviewed history to set the review frequency. When an error is detected, the \emph{recovery path} guides local repair under continued review. After the reviewer verifies recovery, the harness resumes ordinary scheduling. It retains the failure record to guide future interventions.

We evaluate \fh{} on four simulated manipulation tasks, examining both execution efficiency and task success. Across repeated runs, reviewer token use decreases by an average of \MeanTokenReduction{} per task, and mean episode time falls from 85.5 to 52.0 minutes. These comparisons use the first evaluated episode and the mean of the final five episodes. On these same tasks, mean success increases from \HybridMeanSuccessRate{} for the baseline to \SharedTaskMeanSuccessRate{} with \fh{}. Execution traces illustrate how the harness adjusts review frequency and resumes routine execution after recovery.

\section{Related Work}
\label{sec:related}

\paragraph{Language-model-guided robot execution.}
Language models support robot execution through skill selection, planning, and feedback. SayCan grounds skill selection in robotic affordances \citep{ahn2022can}, while Inner Monologue uses execution feedback to guide language-based planning \citep{huang2022inner}. VLA policies such as \pizero{} map visual observations and instructions to robot actions \citep{intelligence2025pi_}. GPT-as-Policy combines \pizero{} with GPT-6 Astra for execution review and correction \citep{su2026astra}. \fh{} builds on this hybrid execution framework, using accumulated experience to schedule review and coordinate recovery.

\paragraph{Selective supervision and intervention.}
Selective supervision concerns when an autonomous system should request external assistance. \citet{ren2023robots} study uncertainty alignment in language-model planners that ask for help. In interactive imitation learning, HG-DAgger incorporates human interventions into policy learning \citep{kelly2019hg}, while ThriftyDAgger uses novelty and risk to manage an intervention budget \citep{hoque2021thriftydagger}. \fh{} addresses this scheduling question during repeated execution with a fixed policy, using reviewed outcomes from recurring task stages to adjust the frequency of model supervision.

\paragraph{Experience and failure recovery.}
Past interactions can inform subsequent planning and error correction. Reflexion uses verbal feedback to guide later trials \citep{shinn2023reflexion}, and Voyager accumulates reusable skills through open-ended interaction \citep{wang2023voyager}. REFLECT summarizes robot experiences to explain failures and support correction \citep{liu2023reflect}. \fh{} uses reviewed experience to allocate supervision across routine execution and recovery, retaining failure evidence after ordinary scheduling resumes.

\section{Problem Setting}
\label{sec:setting}

We consider a robot executing a manipulation task specified by a language instruction. At each decision step, a pretrained vision-language-action (VLA) policy receives the current observation and proposes an action chunk comprising multiple control steps. A separate model-based reviewer can assess recent execution and the proposed actions, then approve the proposal or provide corrective feedback.

The task is executed over a sequence of episodes with varying scene layouts. Within each episode, the harness maintains a history of observations, actions, and feedback. Across episodes, it retains task-specific experience that can inform later decisions. We refer to episodes connected by this persistent memory as a \emph{memory chain}. The VLA policy weights remain fixed throughout the chain; adaptation occurs in the harness's memory and supervision schedule.

The harness determines when to request review and how to incorporate feedback into subsequent execution, including error recovery. We study how accumulated task experience can reduce model-review overhead while maintaining task performance. Task success is defined by the environment's completion criterion.

\section{FastHarness}
\label{sec:method}

\fh{} augments the GPT-as-Policy execution loop with a scheduler that uses task memory to coordinate \pizero{} and GPT-6 Astra (Figure~\ref{fig:system}). For each new observation, the policy proposes an action chunk and the scheduler decides whether it requires review. The fast path adjusts this decision using stage-specific experience; the recovery path maintains review while a detected error is corrected. Reviewed outcomes update the memory used for subsequent decisions.

\subsection{Experience-Driven Fast Path}
\label{sec:fast-path}

\paragraph{Stage recognition and outcome memory.}
The reviewer assigns stage labels, such as approaching, grasping, and placing, during reviewed execution. \fh{} represents observations using downsampled RGB views and proprioceptive state, and stores examples under these labels. It matches each new observation to the nearest stored example for each stage; a distant or ambiguous match triggers review. Stage statistics use only explicit success or failure judgments for the immediately preceding ordinary policy transition, attributed to its pre-action stage. Earlier chunks between reviews receive no individual labels, and corrective or recovery actions are excluded from these counts.

\paragraph{Experience-driven review scheduling.}
For each stage, the scheduler computes a conservative score from reviewed success and failure counts. It combines this score with the amount of reviewed experience to set the interval between reviews. Limited evidence, or insufficient consecutive successes after an error, keeps the interval at one chunk. Higher scores and more reviewed samples permit longer intervals. Unknown or ambiguous contexts, uncertainty from the previous review, execution alerts, and proximity to remembered errors override this interval and trigger review. Stage-transition checks, random audits, and a control-step limit between reviews provide additional review triggers. Appendix~\ref{app:implementation} specifies the matching rule, score, interval function, and trigger priority.

\paragraph{Observation-conditioned execution.}
When no review is required, the executor runs a prefix of the current policy proposal within the remaining control-step allowance since the last review. It then obtains a new observation and proposal before the next scheduling decision. Longer review intervals therefore retain observation-conditioned control between chunks; stored trajectories are not replayed.

\begin{figure*}[!t]
\centering
\begin{subfigure}[t]{.242\textwidth}\centering
\includegraphics[width=\linewidth]{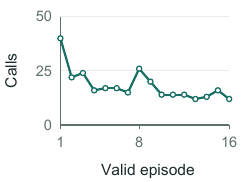}
\caption{Build tower}
\label{fig:panel:review_build_tower}
\end{subfigure}\hfill
\begin{subfigure}[t]{.242\textwidth}\centering
\includegraphics[width=\linewidth]{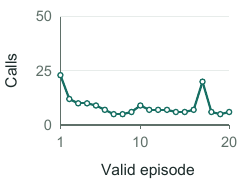}
\caption{Fold clothes}
\label{fig:panel:review_fold_clothes}
\end{subfigure}\hfill
\begin{subfigure}[t]{.242\textwidth}\centering
\includegraphics[width=\linewidth]{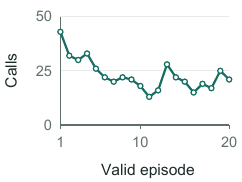}
\caption{Bottle disposal}
\label{fig:panel:review_put_bottles_into_dustbin}
\end{subfigure}\hfill
\begin{subfigure}[t]{.242\textwidth}\centering
\includegraphics[width=\linewidth]{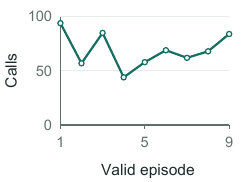}
\caption{Classify objects}
\label{fig:panel:review_classify_objects}
\end{subfigure}
\caption{GPT-6 calls across repeated task executions. Each panel uses a task-specific call-count scale and spans all evaluated episodes: 16 for Build tower, 20 each for Fold clothes and Bottle disposal, and nine for Classify objects. Episode indices follow the evaluation order within each shared-memory chain.}
\label{fig:efficiency}
\end{figure*}

\begin{figure*}[!t]
\centering
\begin{subfigure}[t]{.49\textwidth}\centering
\includegraphics[width=\linewidth]{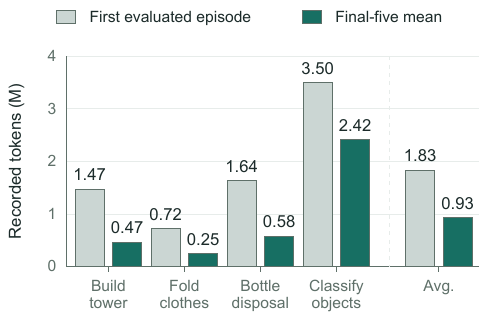}
\caption{Recorded GPT-6 tokens}
\label{fig:tokens}
\end{subfigure}\hfill
\begin{subfigure}[t]{.49\textwidth}\centering
\includegraphics[width=\linewidth]{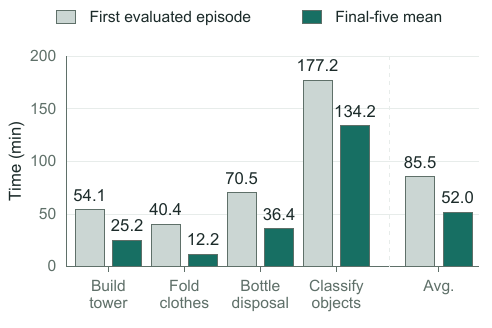}
\caption{Execution time}
\label{fig:runtime}
\end{subfigure}
\caption{Execution efficiency for the first evaluated episode and the final-five-episode mean. Avg. denotes the equal-task mean. (a) Recorded input and output tokens, with cached input counted once. The mean per-task reduction is \MeanTokenReduction{}, and the equal-task mean falls from 1.83 to 0.93 million tokens. (b) Episode time in minutes; the equal-task mean falls from 85.5 to 52.0 minutes (39\% lower).}
\label{fig:execution-efficiency}
\end{figure*}

\subsection{Error Recovery Path}
\label{sec:recovery-path}

\paragraph{Detecting errors and repairing locally.}
A reviewed execution error or intent mismatch activates the recovery path. Every action chunk is reviewed while recovery is active. The reviewer first guides local corrections from the current state, such as re-approaching and re-grasping a dropped object.

\paragraph{Recovering toward verified waypoints.}
If local repair does not resolve the fault, the harness can select a waypoint verified earlier in the episode. It first considers the most recent eligible waypoint and can select an earlier one if needed. The policy generates new actions toward the selected goal in the current environment. Selecting a waypoint does not reset the environment.

\paragraph{Verifying recovery and resuming execution.}
Ordinary scheduling resumes only after the reviewer confirms that the fault has been resolved. For waypoint-based recovery, verifying the return goal is followed by a separate assessment of the original fault. Recovery completion clears the active recovery state while retaining the error record and its recovery evidence. Later observations near that error still require review, even when the stage's overall success record has improved.

\section{Experiments}
\label{sec:results}

\subsection{Experimental Setup}
\label{sec:evaluation}

We evaluate \fh{} in RoboDojo simulated manipulation using \pizero{} as the student policy and GPT-6 Astra as the reviewer. Each task is executed repeatedly with varying scene layouts and persistent task memory. Both efficiency and success are evaluated on four tasks: Build tower, Fold clothes, Bottle disposal, and Classify objects. Baseline success rates for these tasks are taken from the GPT-as-Policy report \citep{su2026astra}.

We measure reviewer calls, review-skip rate (the fraction of completed chunks executed automatically), recorded tokens, and wall-clock time. Unless otherwise specified, efficiency comparisons use the first evaluated episode and the mean of the final five episodes within each task's memory chain. Success follows the native task evaluator. Appendix~\ref{app:experiment-details} provides configurations and metric definitions; Appendices~\ref{app:detailed-results} and~\ref{app:tables} give detailed traces and complete numerical results.

\subsection{Reviewer Use and Execution Efficiency}
\label{sec:efficiency-results}

Reviewer demand decreases between the initial and later evaluation windows (Figure~\ref{fig:efficiency}). In Build tower, the call count falls from 40 to a final-five mean of 13.4, with approximately 75\% of chunks executed automatically. Classify objects retains greater review demand: calls decrease from 94 to 68.2, while approximately 24\% of chunks execute automatically in the final five episodes. The traces show task-dependent reductions in supervision, with review remaining active throughout later execution.

Lower review demand is accompanied by reduced token consumption (Figure~\ref{fig:tokens}). Recorded GPT-6 token use decreases by \MeanTokenReduction{} on average across tasks, computed as the mean of the four per-task percentage reductions. The equal-task mean token count falls from 1.83 million to 0.93 million per episode.

\afterpage{%
\begin{figure*}[!t]
\centering
\includegraphics[width=.95\textwidth]{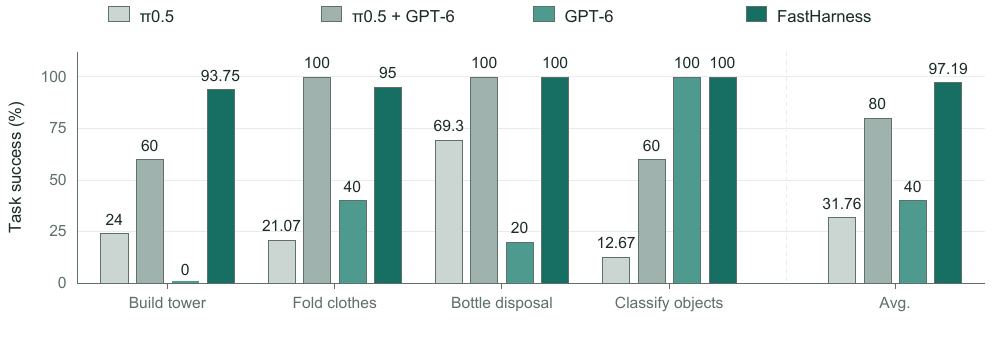}
\caption{Task success rates on the four shared tasks. The mean is \HybridMeanSuccessRate{} for \pizero{} + GPT-6 and \SharedTaskMeanSuccessRate{} for \fh{}. Values are percentages; Avg. denotes the equal-task mean. FastHarness counts are in Table~\ref{tab:success}; the other methods\textquotesingle{} results come from the GPT-as-Policy report \citep{su2026astra}. Appendix~\ref{app:tables} gives the values and their sources.}
\label{fig:success}
\end{figure*}
\begin{figure*}[!t]
\centering
\begin{minipage}{.95\textwidth}
\centering
\begingroup
\fontsize{8}{10}\selectfont\sffamily
\makebox[.6615\linewidth][c]{%
\textcolor[HTML]{CBD6D3}{\rule{9pt}{6pt}}\hspace{4pt}Episodes 1--5\hspace{16pt}%
\textcolor{BlogTeal}{\rule{9pt}{6pt}}\hspace{4pt}Episodes 16--20}%
\hfill\makebox[.323\linewidth][c]{Episodes 1--20}\par
\endgroup
\vspace{3pt}
\begin{subfigure}[t]{.323\linewidth}\centering
\includegraphics[width=\linewidth,trim=0 0 0 16bp,clip]{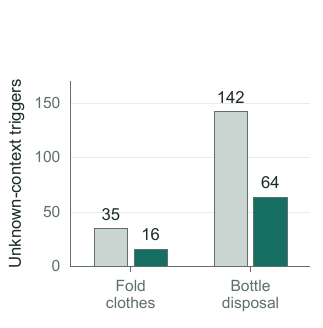}
\caption{Unknown-context triggers}
\label{fig:panel:unknown_context}
\end{subfigure}\hfill
\begin{subfigure}[t]{.323\linewidth}\centering
\includegraphics[width=\linewidth,trim=0 0 0 16bp,clip]{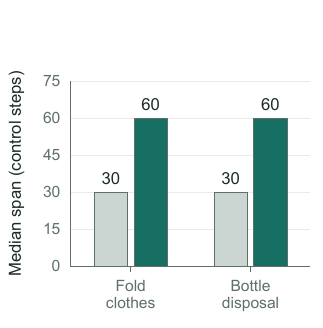}
\caption{Median automatic span}
\label{fig:panel:automatic_span_median}
\end{subfigure}\hfill
\begin{subfigure}[t]{.323\linewidth}\centering
\includegraphics[width=\linewidth,trim=0 0 0 16bp,clip]{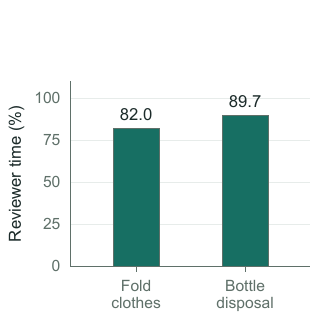}
\caption{Reviewer time share}
\label{fig:cost-breakdown}
\end{subfigure}
\end{minipage}
\caption{Fast-path behavior and reviewer cost in Fold clothes and Bottle disposal. (a) Total unknown-context review triggers in episodes 1--5 and 16--20, excluding terminal audits. (b) Median uninterrupted automatic execution length in the same windows, measured in control steps. In (a,b), lighter and darker bars denote the early and late windows. (c) Reviewer interaction as a percentage of total wall-clock time over all 20 episodes, including request handling, waiting, and retries.}
\label{fig:fast-behavior}
\end{figure*}
}

End-to-end execution also becomes faster (Figure~\ref{fig:runtime}). The equal-task mean episode time falls from 85.5 to 52.0 minutes, a reduction of approximately 39\%. This measure includes policy inference, robot execution, reviewer service time, and runtime overhead.

\subsection{Task Completion}
\label{sec:completion-results}

Across the four shared tasks, \fh{} achieves a mean success rate of \SharedTaskMeanSuccessRate{}, compared with \HybridMeanSuccessRate{} for the hybrid \pizero{} + GPT-6 Astra baseline, a difference of 17.19 percentage points (Figure~\ref{fig:success}). Baseline results are from the GPT-as-Policy report \citep{su2026astra}.

The gains come from Build tower, which improves from 60\% to 93.75\% (15/16), and Classify objects, which improves from 60\% to 100\% (9/9). Bottle disposal remains at 100\% for both methods, with \fh{} completing 20/20 episodes. Fold clothes reaches 95\% (19/20), compared with 100\% for the hybrid. Table~\ref{tab:success} reports the completion counts for all four tasks.

\subsection{Review Allocation and Runtime Analysis}
\label{sec:fast-analysis}

\afterpage{%
\afterpage{%
\begin{figure*}[!t]
\centering
\begin{subfigure}[t]{.192\textwidth}
\centering
\includegraphics[width=\linewidth]{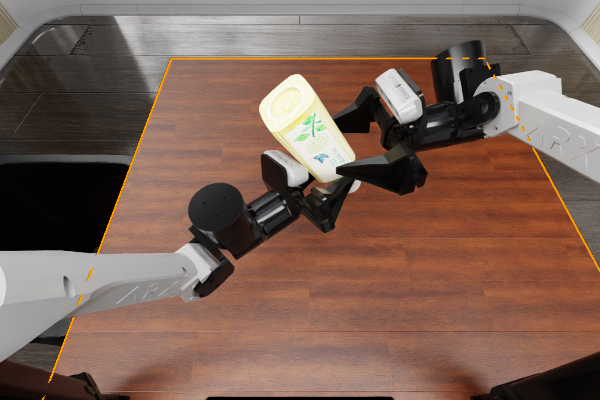}
\caption{Carry}
\label{fig:panel:bottle_frame_1}
\end{subfigure}\hfill
\begin{subfigure}[t]{.192\textwidth}
\centering
\includegraphics[width=\linewidth]{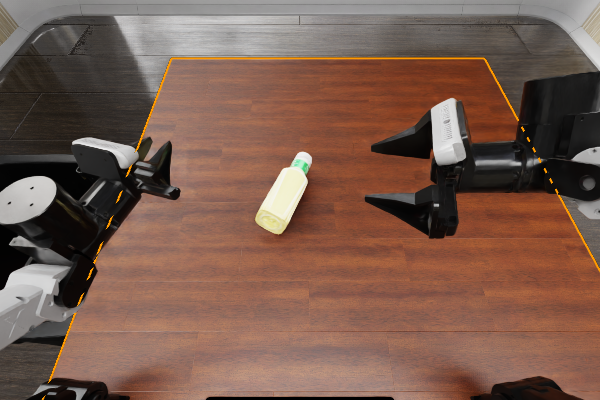}
\caption{Drop}
\label{fig:panel:bottle_frame_2}
\end{subfigure}\hfill
\begin{subfigure}[t]{.192\textwidth}
\centering
\includegraphics[width=\linewidth]{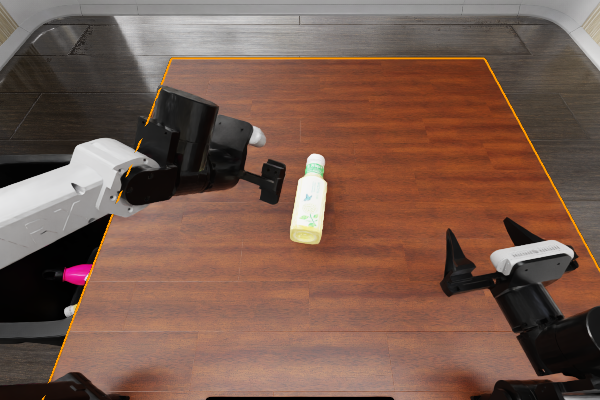}
\caption{Approach}
\label{fig:panel:bottle_frame_3}
\end{subfigure}\hfill
\begin{subfigure}[t]{.192\textwidth}
\centering
\includegraphics[width=\linewidth]{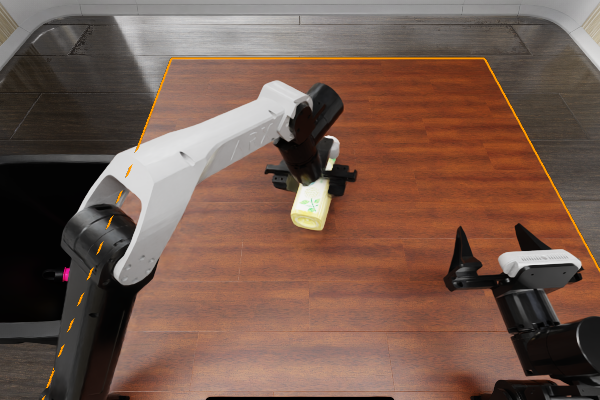}
\caption{Re-grasp}
\label{fig:panel:bottle_frame_4}
\end{subfigure}\hfill
\begin{subfigure}[t]{.192\textwidth}
\centering
\includegraphics[width=\linewidth]{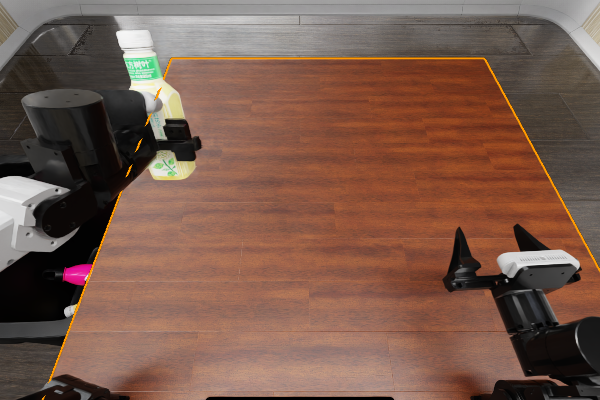}
\caption{Resume}
\label{fig:panel:bottle_frame_5}
\end{subfigure}
\caption{Local repair in Bottle disposal, episode 19. A bottle drops during transport; the robot approaches it again, re-grasps it, and resumes transport. Frames (a)--(e) show control steps 444, 459, 474, 509, and 524 from the same camera. Recovery ends at step 509; automatic execution resumes at 534. The episode finishes with a score of 100/100.}
\label{fig:recovery-frames}
\end{figure*}
\begin{figure*}[!t]
\centering
\begin{minipage}{\textwidth}
\centering
\begin{subfigure}[t]{.192\linewidth}
\centering
\includegraphics[width=\linewidth]{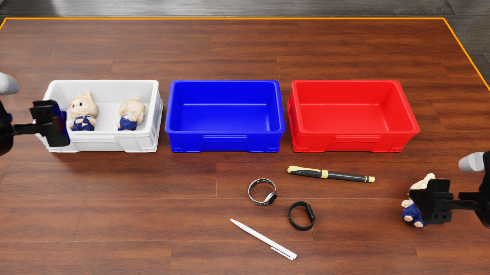}
\caption{Save subgoal\\{\scriptsize Step 174}}
\label{fig:panel:classify_frame_174}
\end{subfigure}\hfill
\begin{subfigure}[t]{.192\linewidth}
\centering
\includegraphics[width=\linewidth]{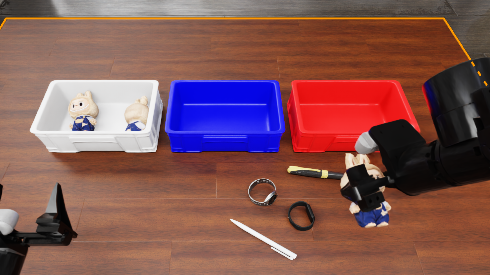}
\caption{Detect mismatch\\{\scriptsize Step 219}}
\label{fig:panel:classify_frame_219}
\end{subfigure}\hfill
\begin{subfigure}[t]{.192\linewidth}
\centering
\includegraphics[width=\linewidth]{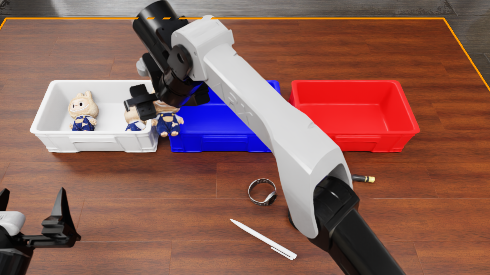}
\caption{Retrieve waypoint\\{\scriptsize Step 299}}
\label{fig:panel:classify_frame_299}
\end{subfigure}\hfill
\begin{subfigure}[t]{.192\linewidth}
\centering
\includegraphics[width=\linewidth]{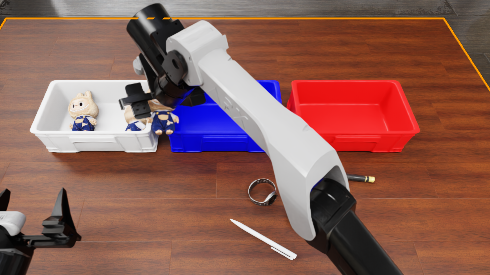}
\caption{Verify subgoal\\{\scriptsize Step 304}}
\label{fig:panel:classify_frame_304}
\end{subfigure}\hfill
\begin{subfigure}[t]{.192\linewidth}
\centering
\includegraphics[width=\linewidth]{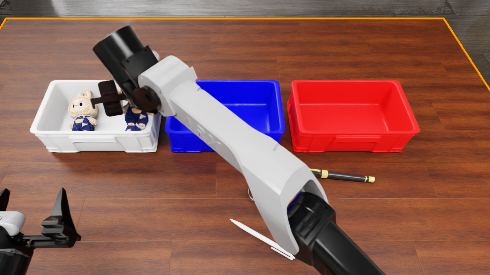}
\caption{Resume execution\\{\scriptsize Step 344}}
\label{fig:panel:classify_frame_344}
\end{subfigure}
\par\vspace{2pt}
\begin{subfigure}[t]{\linewidth}
\centering
\includegraphics[width=\linewidth]{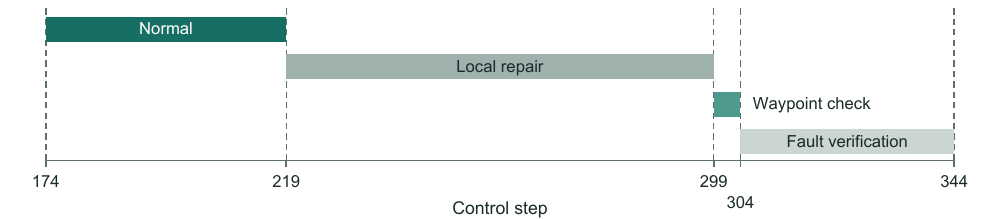}
\caption{Recovery phases aligned with the five observations}
\label{fig:panel:classify_strip_timeline}
\end{subfigure}
\end{minipage}
\caption{Waypoint-guided recovery in the Classify objects task. Frames (a)--(e) show control steps 174, 219, 299, 304, and 344 from the same camera and crop. At step 219, the next proposal would release the third rabbit outside the basket. The harness selects the saved subgoal at 299, verifies that it still holds at 304, and continues correction until normal scheduling resumes at 344. (f) Bar lengths are proportional to the logged control-step intervals.}
\label{fig:recovery-behavior}
\end{figure*}
}%
}

To examine how the fast path changes review allocation and where runtime is spent, we analyze the complete 20-episode chains of Fold clothes and Bottle disposal. For review allocation, we compare episodes 1--5 with episodes 16--20 (Figure~\ref{fig:fast-behavior}a--b).

An \emph{unknown-context trigger} requests review when memory contains no sufficiently matching example of the current context. The total number of these triggers across each five-episode window decreases from 35 to 16 for Fold clothes and from 142 to 64 for Bottle disposal. Thus, fewer review requests arise from unfamiliar contexts in later executions.

An \emph{automatic span} comprises consecutive completed action chunks with no intervening GPT review. Within a span, \pizero{} continues to generate actions from fresh observations between chunks. The median over each window's spans increases from 30 to 60 control steps in both tasks, corresponding to two versus four action chunks. This measures the actual length of uninterrupted automatic execution. The full span distributions appear in Appendix~\ref{app:fast-path-details}, Figure~\ref{fig:automatic-span-distributions}.

Reviewer interaction accounts for 82.0\% of total wall-clock time in Fold clothes and 89.7\% in Bottle disposal, aggregated over all 20 episodes per task (Figure~\ref{fig:cost-breakdown}). This timer includes request handling, waiting, and retries; memory operations contribute less than 0.1\% in each chain. The concentration of runtime in reviewer interaction makes review frequency a useful optimization target. Appendix~\ref{app:fast-path-details}, Figure~\ref{fig:cumulative-usage}, additionally reports running mean token usage, including the initial cost of acquiring experience.

\subsection{Recovery Behavior}
\label{sec:recovery-results}

We examine local repair and correction guided by a verified subgoal. The harness reviews each action chunk until the fault is resolved, then restores ordinary scheduling.

\paragraph{Local repair after a dropped object.}
In the Bottle disposal task, a bottle can slip from the gripper during transport. Figure~\ref{fig:recovery-frames} illustrates the resulting local repair. In this example, recovery starts at control step 459. Under continuous review, the robot approaches the bottle again and re-grasps it. Normal scheduling resumes at step 509, and automatic execution follows at 534. The episode finishes with a score of 100/100. This sequence shows how a local correction allows execution to continue from the current scene.

\paragraph{Using an earlier subgoal to guide correction.}
In one execution of the Classify objects task, the reviewer detects at step 219 that the next proposed action would release the third rabbit outside the basket (Figure~\ref{fig:recovery-behavior}). After local corrections, the harness selects a saved waypoint at step 299. The waypoint records an earlier subgoal: two rabbits placed in the white basket. At step 304, the reviewer confirms that this subgoal is still satisfied. This check establishes which progress has been preserved; the original fault still requires correction. Reviewed execution continues until normal scheduling resumes at step 344. The episode ultimately succeeds after a separate later local repair. Verifying the saved subgoal and resolving the triggering fault are separate decisions.

\paragraph{Observed recovery outcomes.}
All 16 recovery events in Classify episodes 1--8 return to normal scheduling. This measures completion of an intervention, rather than native task success. Detection can still miss task-level failures: in one execution of the Fold clothes task, the harness performs 20 reviews without initiating recovery, yet the task ends in failure with a native score of 0.2. Appendix~\ref{app:detailed-results} provides the event accounting.

\section{Conclusion}

We presented \fh{}, an experience-driven execution harness that adaptively allocates model supervision in agentic robot systems. Its fast path uses accumulated stage experience to reduce redundant review, while its recovery path guides error correction and restores routine execution after recovery is verified. Experiments on four simulated manipulation tasks show that review demand and execution time decrease as experience accumulates. \fh{} achieves a mean task success rate of \SharedTaskMeanSuccessRate{}, compared with \HybridMeanSuccessRate{} reported for the hybrid GPT-as-Policy baseline. These findings highlight adaptive supervision as a practical approach to improving the efficiency and reliability of agentic robot systems without additional model training.

\par
\flushcolsend
\clearpage
\raggedcolsend
\par
\begingroup
\small
\bibliography{references}

@article{ren2023robots,
  title={Robots that ask for help: Uncertainty alignment for large language model planners},
  author={Ren, Allen Z and Dixit, Anushri and Bodrova, Alexandra and Singh, Sumeet and Tu, Stephen and Brown, Noah and Xu, Peng and Takayama, Leila and Xia, Fei and Varley, Jake and others},
  journal={arXiv preprint arXiv:2307.01928},
  year={2023}
}

@article{hoque2021thriftydagger,
  title={{ThriftyDAgger}: Budget-aware novelty and risk gating for interactive imitation learning},
  author={Hoque, Ryan and Balakrishna, Ashwin and Novoseller, Ellen and Wilcox, Albert and Brown, Daniel S and Goldberg, Ken},
  journal={arXiv preprint arXiv:2109.08273},
  year={2021}
}

@inproceedings{kelly2019hg,
  title={{HG-DAgger}: Interactive imitation learning with human experts},
  author={Kelly, Michael and Sidrane, Chelsea and Driggs-Campbell, Katherine and Kochenderfer, Mykel J},
  booktitle={2019 International Conference on Robotics and Automation (ICRA)},
  pages={8077--8083},
  year={2019},
  organization={IEEE}
}

@article{liu2023reflect,
  title={{REFLECT}: Summarizing robot experiences for failure explanation and correction},
  author={Liu, Zeyi and Bahety, Arpit and Song, Shuran},
  journal={arXiv preprint arXiv:2306.15724},
  year={2023}
}

@article{huang2022inner,
  title={{Inner Monologue}: Embodied reasoning through planning with language models},
  author={Huang, Wenlong and Xia, Fei and Xiao, Ted and Chan, Harris and Liang, Jacky and Florence, Pete and Zeng, Andy and Tompson, Jonathan and Mordatch, Igor and Chebotar, Yevgen and others},
  journal={arXiv preprint arXiv:2207.05608},
  year={2022}
}

@article{shinn2023reflexion,
  title={{Reflexion}: Language agents with verbal reinforcement learning},
  author={Shinn, Noah and Cassano, Federico and Gopinath, Ashwin and Narasimhan, Karthik and Yao, Shunyu},
  journal={Advances in neural information processing systems},
  volume={36},
  pages={8634--8652},
  year={2023}
}

@article{wang2023voyager,
  title={{Voyager}: An open-ended embodied agent with large language models},
  author={Wang, Guanzhi and Xie, Yuqi and Jiang, Yunfan and Mandlekar, Ajay and Xiao, Chaowei and Zhu, Yuke and Fan, Linxi and Anandkumar, Anima},
  journal={arXiv preprint arXiv:2305.16291},
  year={2023}
}

@article{ahn2022can,
  title={Do as {I} can, not as {I} say: Grounding language in robotic affordances},
  author={Ahn, Michael and Brohan, Anthony and Brown, Noah and Chebotar, Yevgen and Cortes, Omar and David, Byron and Finn, Chelsea and Fu, Chuyuan and Gopalakrishnan, Keerthana and Hausman, Karol and others},
  journal={arXiv preprint arXiv:2204.01691},
  year={2022}
}

@article{intelligence2025pi_,
  title={{$\pi_{0.5}$}: A Vision-Language-Action Model with Open-World Generalization},
  author={{Physical Intelligence} and Black, Kevin and Brown, Noah and Darpinian, James and Dhabalia, Karan and Driess, Danny and Esmail, Adnan and Equi, Michael and Finn, Chelsea and Fusai, Niccolo and others},
  journal={arXiv preprint arXiv:2504.16054},
  year={2025}
}

@misc{su2026astra,
  title        = {{GPT 6 Astra} as an Embodied Policy},
  author       = {Su, Jiayi and Zheng, Yixin and Yan, Mi and Yi, Li and Zhang, Zhizheng and Wang, He},
  year         = {2026},
  howpublished = {Technical report and code},
  url          = {https://github.com/anonymous-report-421/eval-of-gpt-6-astra-as-policy}
}

@article{zhang2026harness,
  title={Harness vla: Steering frozen vlas into reliable manipulation primitives via memory-guided agents},
  author={Zhang, Yixian and Zhang, Huanming and Gao, Feng and Li, Xiao and Liu, Zhihao and Nie, Yi and Zhu, Chunyang and Qiu, Jiaxing and Yan, Yuchen and Liu, Jiyuan and others},
  journal={arXiv preprint arXiv:2607.08448},
  year={2026}
}

@article{fu2026cap,
  title={CaP-X: A framework for benchmarking and improving coding agents for robot manipulation},
  author={Fu, Letian and Yu, Justin and El-Refai, Karim and Kou, Ethan and Xue, Haoru and Huang, Huang and Xiao, Wenli and Wang, Guanzhi and Niu, Dantong and Li, Fei-Fei and others},
  journal={arXiv preprint arXiv:2603.22435},
  year={2026}
}

@article{lu2026aspire,
  title={ASPIRE: Agentic/Skills Discovery for Robotics},
  author={Lu, Runyu and Wu, Yubo and Kou, Ethan and Fu, Letian and Xiao, Wenli and Mandlekar, Ajay and Xu, Yinzhen and Shi, Guanya and Goldberg, Ken and Chen, Ang and others},
  journal={arXiv preprint arXiv:2607.00272},
  year={2026}
}

@article{shi2025hi,
  title={Hi robot: Open-ended instruction following with hierarchical vision-language-action models},
  author={Shi, Lucy Xiaoyang and Ichter, Brian and Equi, Michael and Ke, Liyiming and Pertsch, Karl and Vuong, Quan and Tanner, James and Walling, Anna and Wang, Haohuan and Fusai, Niccolo and others},
  journal={arXiv preprint arXiv:2502.19417},
  year={2025}
}

@article{kim2024openvla,
  title={Openvla: An open-source vision-language-action model},
  author={Kim, Moo Jin and Pertsch, Karl and Karamcheti, Siddharth and Xiao, Ted and Balakrishna, Ashwin and Nair, Suraj and Rafailov, Rafael and Foster, Ethan and Lam, Grace and Sanketi, Pannag and others},
  journal={arXiv preprint arXiv:2406.09246},
  year={2024}
}

@article{black2024pi_0,
  title={$pi\_0 $: A Vision-Language-Action Flow Model for General Robot Control},
  author={Black, Kevin and Brown, Noah and Driess, Danny and Esmail, Adnan and Equi, Michael and Finn, Chelsea and Fusai, Niccolo and Groom, Lachy and Hausman, Karol and Ichter, Brian and others},
  journal={arXiv preprint arXiv:2410.24164},
  year={2024}
}

@article{lin2025evo,
  title={Evo-0: Vision-language-action model with implicit spatial understanding},
  author={Lin, Tao and Li, Gen and Zhong, Yilei and Zou, Yanwen and Du, Yuxin and Liu, Jiting and Gu, Encheng and Zhao, Bo},
  journal={arXiv preprint arXiv:2507.00416},
  year={2025}
}

@inproceedings{lin2026evo,
  title={Evo-1: Lightweight vision-language-action model with preserved semantic alignment},
  author={Lin, Tao and Zhong, Yilei and Du, Yuxin and Zhang, Jingjing and Liu, Jiting and Chen, Yinxinyu and Gu, Encheng and Liu, Ziyan and Cai, Hongyi and Zou, Yanwen and others},
  booktitle={Proceedings of the IEEE/CVF conference on computer vision and pattern recognition},
  pages={13397--13406},
  year={2026}
}

@article{lin2026evodepth,
  title={Evo-depth: A lightweight depth-enhanced vision-language-action model},
  author={Lin, Tao and Du, Yuxin and Liu, Jiting and Zhu, Nuobei and Li, Yunhe and Fu, Yuqian and Chen, Yinxinyu and Cai, Hongyi and Ye, Zewei and Cheng, Bing and others},
  journal={arXiv preprint arXiv:2605.14950},
  year={2026}
}

@article{brohan2023rt,
  title={Rt-2: Vision-language-action models transfer web knowledge to robotic control},
  author={Brohan, Anthony and Brown, Noah and Carbajal, Justice and Chebotar, Yevgen and Chen, Xi and Choromanski, Krzysztof and Ding, Tianli and Driess, Danny and Dubey, Avinava and Finn, Chelsea and others},
  journal={arXiv preprint arXiv:2307.15818},
  year={2023}
}

@article{lin2026la4vla,
  title={LA4VLA: Learning to Act without Seeing via Language-Action Pretraining},
  author={Lin, Tao and Du, Yuxin and Mao, Yiran and Ye, Zewei and Zhong, Yilei and Cheng, Bing and Wang, Yiming and Liu, Jiting and Tian, Yang and Yan, Junchi and others},
  journal={arXiv preprint arXiv:2606.27295},
  year={2026}
}
\par
\endgroup
\flushcolsend
\clearpage
\raggedcolsend
\appendix
\renewcommand{\UrlFont}{\rmfamily\itshape}
\onecolumn
\small
\setlength{\intextsep}{6pt plus 2pt minus 2pt}
\setlength{\abovecaptionskip}{6pt}
\section{Implementation Details}
\label{app:implementation}
\begingroup
\setlength{\abovedisplayskip}{6pt plus 2pt minus 2pt}
\setlength{\belowdisplayskip}{6pt plus 2pt minus 2pt}

Table~\ref{tab:runtime-config} lists the runtime settings and hyperparameters used throughout this appendix.

\subsection{Stage Recognition and Outcome Attribution}
\label{app:stage-memory}

\paragraph{Stage examples.}
Harness memory matching uses $x\in[0,1]^{590}$: 576 RGB values from three BOX-resized $8\times8$ views normalized by 255, twelve joint angles mapped by $\operatorname{clip}((v+\pi)/(2\pi),0,1)$, and two gripper commands clipped to $[0,1]$. The policy and reviewer retain their respective observation inputs. Each nonterminal review stores the feature under the reviewer's stage label, retaining the latest 128 distinct examples per stage. For stage $s$ with examples $E_s$,
\begin{equation}
d_s(x)=\min_{e\in E_s}\sqrt{\frac{1}{590}\sum_{j=1}^{590}(x_j-e_j)^2}.
\label{eq:stage-distance}
\end{equation}
A context is unknown if memory is empty or the minimum distance exceeds $0.06$, and ambiguous if the two smallest stage distances differ by less than $0.01$. A hash of task, policy configuration, feature representation, reviewer configuration/protocol, and scope partitions memory.

\paragraph{Outcome statistics.}
Each review labels only the immediately preceding transition, using its pre-action feature and stage. Counts include ordinary policy transitions outside recovery labeled \textit{ok} or \textit{error}; \textit{unknown} leaves them unchanged. Re-reviews of one observation attribute the preceding outcome and any intent error once, but update the current stage example each time.

\paragraph{Error memory.}
Separate error records merge when stage and error type agree and feature distance is at most $0.04$. Corrective-action errors and intent mismatches reset the clean streak and mark a past failure without increasing ordinary-policy outcome counts. Verified recovery increments the error's recovery count and stores evidence.

\subsection{Reliability and Review Allocation}
\label{app:review-allocation}

For stage $s$, let $S_s$ and $F_s$ denote reviewed successes and failures, $n_s=S_s+F_s$, and $\hat p_s=S_s/n_s$. The scheduler uses the conservative score
\begin{equation}
r_s=\max\!\left(0,\frac{\hat p_s+\frac{z^2}{2n_s}-z\sqrt{\frac{\hat p_s(1-\hat p_s)}{n_s}+\frac{z^2}{4n_s^2}}}{1+\frac{z^2}{n_s}}\right),
\label{eq:reliability-score}
\end{equation}
with $z=1.96$ and $r_s=0$ when $n_s=0$. Let $T_s$ be the number of labeled ordinary policy transitions, $c_s$ the consecutive successful labeled transitions since the last error, and $f_s$ an indicator of any past error. Here, count decay is $1.0$, so $T_s=n_s$. The review interval is
\begin{equation}
I_s=\begin{cases}
1,&T_s<m\ \text{or}\ (f_s\land c_s<c_{\min}),\\[2pt]
1+\left\lfloor(L-1)r_s^2\min(C,T_s/m)\right\rfloor,&\text{otherwise}.
\end{cases}
\label{eq:review-interval}
\end{equation}
We use $m=2$, $c_{\min}=3$, $L=20$, and $C=6$. Here, $L$ is an interval scale, not a hard cap. Cooldown requires reviewed successes and leaves error-proximity review active.

\paragraph{Review triggers.}
The scheduler selects the first applicable condition in the following priority order: always-review mode, episode start, active recovery, unknown or ambiguous context, optional contact gating, stage change, previous uncertainty, monitor or proposal warning, error proximity, control-step allowance, scheduled interval, and random audit. Contact gating is disabled in the evaluated configuration, and a familiar stage change may proceed without review when $I_s>1$. A same-stage error within distance $0.04$ requires review regardless of its recovery count or the stage's clean streak. Reported trigger counts follow this priority order; monitoring and audit settings appear in Table~\ref{tab:runtime-config}.

\paragraph{Interval and control-step accounting.}
The chunk counter resets before reviewed execution and increments after every executed chunk. A fixed interval $I_s=k$ therefore permits at most $k-1$ subsequent automatic chunks, subject to earlier triggers. The base allowance of 120 control steps starts at the most recent reviewed observation and includes the ensuing execution. Automatic execution uses a fresh proposal prefix within the remaining allowance. Scheduled chunk intervals and realized automatic control-step spans are thus distinct.

\subsection{Control Loop and Recovery State}
\label{app:control-loop}

Scheduling uses memory available before review. Recovery phase changes and attempt completion trigger re-review of the same observation and proposal. An \textit{unknown} outcome or \textit{uncertain} intent forces review at the next decision. Native task completion is evaluated separately. Algorithm~\ref{alg:control-loop} summarizes the control loop.

\paragraph{Recovery progression.}
Reviewed execution errors or intent mismatches initiate local repair. Each attempt ends on an explicit success or failure judgment, or after eight executed chunks without success. Two unsuccessful local attempts trigger correction toward the latest eligible waypoint, if available; an unsuccessful waypoint attempt prompts reviewer selection of another target. Eligible waypoints precede the failure and exclude attempted targets. They are saved outside recovery after an \textit{ok} outcome, with a reviewer-supplied goal and prerequisites. Verified return enters \textit{after\_return}. At least one further chunk must execute before a \textit{succeeded} judgment permits \textit{normal} execution. All successful recovery judgments require the preceding outcome to be \textit{ok}.

\begin{algorithm}[H]
\caption{Review scheduling and recovery before each action chunk in \fh{}.}
\label{alg:control-loop}
\footnotesize

\begin{tabular}{@{}r@{\hspace{0.9em}}p{0.94\linewidth}@{}}
1 & Match the current observation; obtain and validate a fresh policy proposal. \\
2 & Compute the interval and control-step allowance; set the review flag from the ordered triggers. \\
3 & \textbf{while} the review flag is set \textbf{do} \\
4 & \quad Review the observation, proposal, preceding transition, history, memory, and recovery context. \\
5 & \quad On the first review only: attribute outcome and intent error; store any eligible checkpoint. \\
6 & \quad Update the stage example, reset review counters, and update recovery. \\
7 & \quad Retain outcome/intent uncertainty; set the review flag if recovery requests re-review. \\
8 & \textbf{end while} \\
9 & Use the approved action if reviewed; otherwise, select a proposal prefix within the remaining allowance. \\
10 & Execute the selected chunk and validate the acknowledgment. \\
11 & Update history, monitoring, execution spans, and chunk counters. \\
\end{tabular}
\end{algorithm}

\paragraph{Memory example.}
Table~\ref{tab:memory-example} summarizes the final Bottle disposal memory.

\smallskip
\noindent\begin{minipage}{\linewidth}
\centering
\footnotesize
\renewcommand{\arraystretch}{1.08}
\begin{tabular}{@{}p{0.29\linewidth}p{0.67\linewidth}@{}}
\toprule
Memory entry & End-of-chain state \\
\midrule
Reviewed outcomes & 77 successes, 1 failure; 78 labeled ordinary policy transitions \\
Stage history & Clean streak: 9; past-failure indicator: \textit{true} \\
Error record & \textit{dropped\_bottle}; 1 occurrence, 1 verified recovery \\
Visual evidence & Bottle leaves gripper at steps 444--459; verified lift at steps 504--509 \\
\bottomrule
\end{tabular}
\captionsetup{hypcap=false}
\captionof{table}{Final \textit{transport\_bottle} memory; recovered errors remain eligible for proximity review.}
\label{tab:memory-example}
\end{minipage}
\smallskip
\endgroup

\section{Experimental Setup and Measurement}
\label{app:experiment-details}

\subsection{Runtime and Controller Configuration}

We run RoboDojo on NVIDIA H20 hardware with CPU-based physics. The software environment comprises Python 3.11.15, Isaac Sim 5.1.0, JAX/JAXlib 0.5.3, Flax 0.10.2, and NumPy 1.26.4. Table~\ref{tab:runtime-config} specifies the policy, observations, and harness configuration.

\begingroup
\small
\setlength{\tabcolsep}{5pt}
\renewcommand{\arraystretch}{1.05}
\setlength{\LTcapwidth}{\textwidth}
\begin{longtable}{@{}p{0.24\linewidth}p{0.72\linewidth}@{}}
\caption{Environment, policy, and harness configuration used for evaluation.}\label{tab:runtime-config}\\
\toprule
Component & Configuration \\
\midrule
\endfirsthead
\multicolumn{2}{l}{\tablename\ \thetable\ (continued)}\\
\toprule
Component & Configuration \\
\midrule
\endhead
\midrule
\multicolumn{2}{r}{\emph{Continued on next page}}\\
\endfoot
\bottomrule
\endlastfoot
Environment & RoboDojo native runtime on Isaac Sim; two X5 arms; CPU physics. \\
Control and observations & Control step $0.04$\,s (25\,Hz); three RGB views (high, left wrist, right wrist), joint state and end-effector state; no depth input. \\
Policy & \pizero{}, OpenPI/JAX; configuration \textit{pi05\_base\_aloha\_full\_sim\_arx-x5\_seed\_0}, checkpoint directory \textit{59999}; fixed weights. \\
Policy randomness & JAX seed 0; a fresh policy server for each episode. \\
Action interface & 14 dimensions: six joints and one gripper opening per arm; internal arm actions are increments relative to the current joint state; gripper 0 is closed and 1 is open. The policy predicts 50 actions; an executed proposal contains 1--15 control steps. \\
Reviewer & \textit{gpt-6-astra}, reasoning effort \textit{xhigh}; structured review output via the Responses protocol. \\
Memory representation & Three $8\!\times\!8$ RGB thumbnails and 14 normalized state values (590 dimensions) for harness memory matching. The policy and reviewer retain their respective observation inputs. \\
\midrule
Matching & Root-mean-square feature distance; match threshold 0.06, ambiguity margin 0.01, nearby-error radius 0.04. \\
Review interval & Interval scale $L=20$, minimum sample count $m=2$, experience multiplier cap $C=6$, and cooldown $c_{\min}=3$ reviewed successes. The scale is not a hard interval limit. \\
Audits and monitoring & Random audit probability 0.03, seed 0 reset per episode; stationary-state alert after 3 chunks with all state changes below $10^{-5}$; base allowance of 120 control steps since the most recent review. \\
Memory updates & Ordinary policy outcome-count decay 1.0; at most 128 examples per stage; stage-span update $0.7\,\text{old}+0.3\,\text{new}$. Memory is shared across repetitions within each task chain. \\
Enabled mechanisms & Persistent error review and relaxation of familiar stage changes enabled. The analyzed execution spans remain within the 120-control-step base allowance. \\
Optional contact gate & Disabled. \\
Recovery configuration & Checkpoint capacity 64; up to 8 executed chunks per attempt; waypoint-directed correction after 2 unsuccessful local attempts when an eligible waypoint is available. \\
\end{longtable}
\endgroup

\subsection{Evaluation Protocol}

\begin{table}[!htbp]
\centering
\small
\setlength{\tabcolsep}{5pt}
\renewcommand{\arraystretch}{1.10}
\begin{tabular}{@{}p{0.22\linewidth}>{\raggedleft\arraybackslash}p{0.16\linewidth}p{0.56\linewidth}@{}}
\toprule
Task & Native step limit & Evaluation and analysis \\
\midrule
Build tower & 1,050 & 16 episodes: efficiency and completion \\
Fold clothes & 500 & 20 episodes: review allocation, automatic spans, time and tokens \\
Bottle disposal & 700 & 20 episodes: review allocation, automatic spans, time and tokens \\
Classify objects & 1,100 & 9 episodes: efficiency and completion; episodes 1--8: 16 recovery events \\
\bottomrule
\end{tabular}
\caption{Task coverage and native control-step limits. Task success counts are reported in Table~\ref{tab:success}.}
\label{tab:task-protocol}
\end{table}

\paragraph{Task chains.}
The four tasks use RoboDojo's standard panel configuration with varying scene layouts. Memory persists across episodes within each task chain, and efficiency and completion are evaluated on the same episodes. Fold clothes and Bottle disposal each contain 20 episodes; Build tower contains 16 and Classify objects contains nine. Review is forced at the start of Classify objects. Build tower's first evaluated episode follows earlier task experience and corresponds to source episode 2. Figures index evaluated episodes consecutively; the plotting data retain the source-index mapping.

\paragraph{Task outcomes.}
Success is determined by the native evaluator's \textit{native\_success} flag at termination or truncation. Runs interrupted by network or runtime infrastructure faults are outside the evaluation; task failures remain in the denominator. Reviewer approval and partial scores do not alter the native outcome. Each task uses case identifier \textit{<task>\_\_standard\_\_g0\_\_l0} and the control-step limit in Table~\ref{tab:task-protocol}. Reproduction uses the corresponding task assets, native predicates, and geometric tolerances.

\subsection{Metrics and Aggregation}

\paragraph{Review and token usage.}
For $N_{\mathrm{auto}}$ automatic chunks among $N_{\mathrm{all}}$ completed chunks, the review-skip rate is $r_{\mathrm{skip}}=N_{\mathrm{auto}}/N_{\mathrm{all}}$. A review decision may require several service calls during re-planning; terminal audits add calls without executing another chunk. Calls, accepted reviews, and executed chunks are therefore counted separately; the skip rate cannot be inferred as $1-\text{calls}/\text{chunks}$. Token usage is the recorded sum of input and output tokens, with cached input counted once. These are usage counts rather than monetary costs.

\paragraph{Runtime.}
Total wall-clock time includes policy inference, robot execution, reviewer interaction, and runtime overhead. Reviewer time includes request handling, waiting, and retries. The remaining time is reported as \emph{other}, which aggregates all components outside reviewer interaction.

\paragraph{Comparison windows.}
Efficiency compares the first evaluated episode with the mean of the final five episodes in each chain. For Classify objects, this is episode 1 versus episodes 5--9; its recovery-event analysis covers episodes 1--8. Review allocation and automatic-span distributions compare episodes 1--5 with episodes 16--20 in Fold clothes and Bottle disposal.

\paragraph{Automatic spans and cumulative usage.}
An automatic span consists of consecutive completed automatic chunks without an intervening review. We pool spans within each comparison window: the early and late windows contain 21 and 19 spans for Fold clothes, and 18 and 26 for Bottle disposal. Episodes without automatic execution contribute no spans. Trigger counts identify the first applicable scheduling condition and exclude terminal audits. For episode token counts $T_i$, the running mean is $\bar T_n=\sum_{i=1}^{n}T_i/n$, including all evaluated episodes through $n$. Appendix~\ref{app:fast-path-details} reports these distributions and cumulative-usage curves.

Episodes share task memory, and a recovery event may contain several attempts. We therefore report descriptive chain-level results without treating episodes or recovery events as independent trials. Returning to normal scheduling records the end of an intervention; task success remains a separate native outcome.

\section{Recovery Trace Analysis}
\label{app:detailed-results}

This appendix details the recovery traces and event outcomes discussed in Section~\ref{sec:results}. Appendix~\ref{app:experiment-details} defines the configurations and metrics, and Appendix~\ref{app:tables} provides the numerical results.

\paragraph{Local repair and resumed automatic execution.}
In Bottle disposal episode 19 (Figure~\ref{fig:recovery-frames}), local recovery begins at step 459 and ends with a return to normal scheduling at 509; automatic execution resumes at 534. The recovery window contains four completed action chunks and five accepted reviews, consuming 161,222 recorded input-plus-output tokens over 480.15 seconds of harness elapsed time. These measurements exclude the triggering review and the replan after recovery exit. The final memory retains the dropped-bottle record and its failure and recovery evidence after the fault is resolved.

\paragraph{Waypoint-guided correction.}
In Classify objects episode 8, an intent mismatch in the next action proposal initiates recovery at step 219, although the preceding execution is judged successful. At step 299, the harness selects the waypoint saved at 174 and enters waypoint-directed recovery. The harness executes a revised action over steps 299--304. At step 304, the reviewer confirms that the saved subgoal---two rabbits placed in the white basket---is still satisfied. Further correction continues until normal scheduling resumes at 344. This verifies preserved progress in the current scene without resetting the physical state. The episode ultimately succeeds after a separate later local-recovery event.

\begin{table}[!htbp]
\centering
\small
\begin{tabular}{lr}
\toprule
Observed event outcome & Count \\
\midrule
Resumed normal scheduling & 16 \\
Closed without normal resumption & 0 \\
Still active at observation end & 0 \\
\midrule
Total recovery events & 16 \\
\bottomrule
\end{tabular}
\caption{Recovery outcomes in Classify objects episodes 1--8, comprising 16 events across eight episodes. Resuming normal scheduling is an intermediate recovery outcome, distinct from native task success. Events within an episode and shared memory chain are not independent trials.}
\label{tab:recovery-events}
\end{table}

\paragraph{Recovery outcomes and failure detection.}
All 16 recovery events in these eight Classify objects episodes return to normal scheduling (Table~\ref{tab:recovery-events}); an event may include multiple local attempts or waypoint checks. Fold clothes episode 17 illustrates a failure that remains undetected by the recovery trigger: the harness performs 20 reviews without initiating recovery, while the native evaluator reports failure with a score of 0.2. Thus, reviewing individual action segments does not ensure detection of every task-level failure. These observations characterize recovery and detection behavior; they do not estimate outcomes without recovery.

\section{Supplementary Numerical Results}
\label{app:tables}

\begin{table}[H]
\centering
\small
\begin{tabular}{lrrr}
\toprule
Task & First episode (M) & Final five (M) & Reduction \\
\midrule
Build tower & 1.47 & 0.47 & 68\% \\
Fold clothes & 0.72 & 0.25 & 65\% \\
Bottle disposal & 1.64 & 0.58 & 64\% \\
Classify objects & 3.50 & 2.42 & 31\% \\
\bottomrule
\end{tabular}
\caption{Recorded GPT-6 token usage (millions) in the first evaluated episode and the mean of the final five episodes. Counts include input and output tokens, with cached input counted once. Build tower's first evaluated episode follows earlier task experience. Values are rounded; the unweighted mean of the four per-task reductions is \MeanTokenReduction{}.}
\label{tab:tokens}
\end{table}

\begin{table}[H]
\centering
\small
\begin{tabular}{lrrr}
\toprule
Task & Episodes & Successes & Rate (\%) \\
\midrule
Build tower & 16 & 15 & 93.75 \\
Fold clothes & 20 & 19 & 95 \\
Bottle disposal & 20 & 20 & 100 \\
Classify objects & 9 & 9 & 100 \\
\bottomrule
\end{tabular}
\caption{FastHarness completion counts and success rates, with an unweighted task mean of \SharedTaskMeanSuccessRate{}. The evaluation protocol is specified in Appendix~\ref{app:experiment-details}.}
\label{tab:success}
\end{table}

\begin{table}[H]
\centering
\small
\begin{tabular}{lrrrr}
\toprule
External method & Build tower & Fold clothes & Bottle disposal & Classify objects \\
\midrule
\pizero{} & 24 & 21.07 & 69.30 & 12.67 \\
\pizero{}+GPT-6 & 60 & 100 & 100 & 60 \\
GPT-6 only & 0 & 40 & 20 & 100 \\
\bottomrule
\end{tabular}
\caption{Success rates (\%) from the GPT-as-Policy report \citep{su2026astra} for the same four tasks.}
\label{tab:external-success}
\end{table}

\paragraph{Baseline results.}
The GPT-as-Policy report \citep{su2026astra} reweights official \pizero{} results from the RoboDojo leaderboard for its selected tasks and scene mix. Its local hybrid and GPT-only evaluations each use five cases per task: two standard and three random cases for Fold clothes, and five standard cases for each of the other three tasks.

Across these four tasks, the unweighted mean success rate is \HybridMeanSuccessRate{} for the GPT-as-Policy hybrid (\pizero{} + GPT-6 Astra) and \SharedTaskMeanSuccessRate{} for \fh{} (Table~\ref{tab:success}).

\section{Supplementary Fast-Path Analysis}
\label{app:fast-path-details}

\begin{figure}[H]
\centering
\begin{subfigure}[t]{.36\textwidth}\centering
\includegraphics[width=\linewidth]{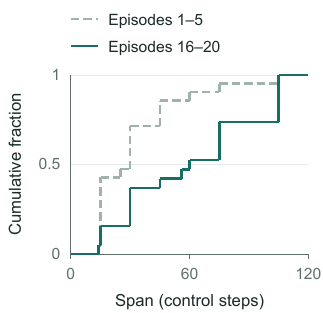}
\caption{Fold clothes}
\label{fig:panel:spans_fold}
\end{subfigure}\hspace{.06\textwidth}
\begin{subfigure}[t]{.36\textwidth}\centering
\includegraphics[width=\linewidth]{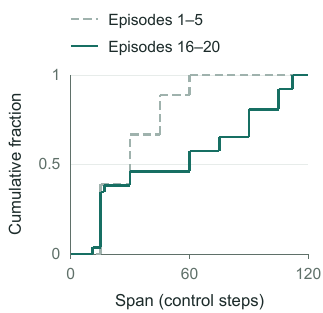}
\caption{Bottle disposal}
\label{fig:panel:spans_bottle}
\end{subfigure}
\caption{Empirical cumulative distributions of automatic-span lengths in episodes 1--5 and 16--20. A rightward shift indicates longer execution spans between GPT-6 reviews.}
\label{fig:automatic-span-distributions}
\end{figure}

\paragraph{Distribution of continuous automatic execution.}
Figure~\ref{fig:automatic-span-distributions} shows empirical cumulative distributions of automatic-span lengths, pooled over episodes 1--5 and 16--20. Each point gives the fraction of spans no longer than the indicated number of control steps. The early and late windows contain 21 and 19 spans for Fold clothes, and 18 and 26 for Bottle disposal. In both tasks, the median increases from 30 to 60 control steps, consistent with the longer automatic execution reported in Figure~\ref{fig:fast-behavior}b. Episodes without automatic execution contribute no spans; Appendix~\ref{app:experiment-details} defines the aggregation procedure.

\begin{figure}[H]
\centering
\begin{subfigure}[t]{.36\textwidth}\centering
\includegraphics[width=\linewidth]{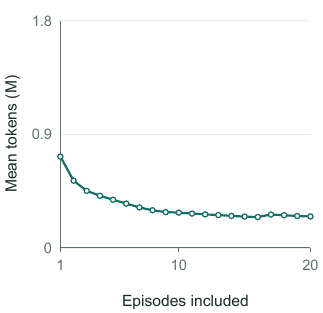}
\caption{Fold clothes}
\label{fig:panel:cumulative_fold}
\end{subfigure}\hspace{.06\textwidth}
\begin{subfigure}[t]{.36\textwidth}\centering
\includegraphics[width=\linewidth]{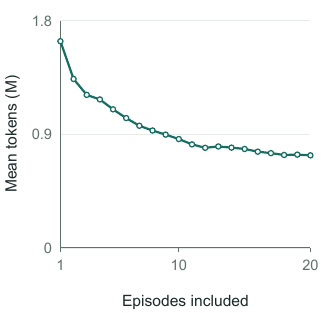}
\caption{Bottle disposal}
\label{fig:panel:cumulative_bottle}
\end{subfigure}
\caption{Running mean recorded tokens per episode on common axes. Each point averages all episodes through the indicated index, including initial experience acquisition.}
\label{fig:cumulative-usage}
\end{figure}

\paragraph{Average token use over repeated execution.}
Figure~\ref{fig:cumulative-usage} reports the running mean token usage, $\bar T_n=\sum_{i=1}^{n}T_i/n$, over the first $n$ episodes. This measure includes the initial cost of acquiring experience. After 20 episodes, the means are 0.249 million tokens for Fold clothes and 0.734 million for Bottle disposal. These chain-level averages complement the first-episode and final-five comparison in Figure~\ref{fig:execution-efficiency} by retaining the cost of all intervening executions.

\section{Limitations}
\label{app:limitations}

Our evaluation covers four simulated RoboDojo tasks using \pizero{} and GPT-6 Astra. The memory representation uses RGB observations and proprioceptive state; deployment with different sensor interfaces may require adapting this representation. Absolute runtime savings also depend on reviewer latency and execution hardware.

\end{document}